\documentclass{ifacconf}
\usepackage{cite}
\usepackage{amsmath,amssymb,amsfonts,graphicx,bbold,mathtools}
\usepackage{enumitem}
\usepackage{subfigure,algorithm}
\usepackage{array}
\usepackage{xcolor}
\usepackage{graphicx}      
\usepackage{natbib}        
\begin{document}
	\begin{frontmatter}
		
		\title{Learning sparse linear dynamic networks in a hyper-parameter free setting }
		
		\thanks[footnoteinfo]{This work was supported by the Wallenberg AI, Autonomous Systems and Software Program (WASP), and the NewLEADS (New directions in learning dynamical systems) project (Contract 2016-06079)}
		
		\author[First]{Arun Venkitaraman} 
		\author[First]{H\aa kan Hjalmarsson} 
		\author[First]{Bo Wahlberg}
		
		\address[First]{Divison of Decision and Control Systems, KTH Royal Institute of Technology, Stockholm, Sweden. \\
			(e-mail: \{arunv,hjalmars,bo\}@ kth.se).}

		\begin{abstract}                
			We address the issue of estimating the topology and dynamics of sparse linear dynamic networks in a hyperparameter-free setting. We propose a method to estimate the network dynamics in a computationally efficient and parameter tuning-free iterative framework known as SPICE (Sparse Iterative Covariance Estimation). The estimated dynamics directly reveal the underlying topology. Our approach does not assume that the network is undirected and is applicable even with varying noise levels across the modules of the network. We also do not assume any explicit prior knowledge on the network dynamics. Numerical experiments with realistic dynamic networks illustrate the usefulness of our method.
		\end{abstract}
		
		
	\end{frontmatter}
	
	\section{Introduction}
	Estimation of dynamic networks have been of increasing interest in the systems and control community. Networks are used to efficiently model the signal flow or the dependencies between the different interconnected modules in a control system \citep{5406168}. Given the set of signals which jointly evolve over time across the nodes of a network, it is often of interest to estimate how they influence each other \citep{Chiuso2012}. In other words, one wishes to estimate the underlying topology and dynamics of the network.
	
	In the case of linear dynamic networks with $J$ nodes or modules, the evolution of the signal $w_i$ at the $i$th node over time $t$ is modelled as \citep{DBLP:journals/corr/abs-1903-06205,VANDENHOF201823}
	\begin{align}
	w_i(t)=\sum_{j=1, j\neq i}G_{ij}(q)w_j(t)+H_{i}(q)e_i(t),
	\label{eq:dyn_sys_model_1}
	\end{align}
	where $G_{ij}$ denotes the transfer function of the module connecting the $j$th node to the $i$th node, $H_i$ denotes the transfer function for the noise or innovation $e_i(t)$ at the $i$th node with variance $\sigma_i^2$, and $q^{-1}$ denotes the unit delay operation. By concatenating the signal equations at all the $J$ nodes, we have the complete model given by
	\begin{align*}
	\mathbf{w}(t)=\mathbf{G}(q)\mathbf{w}(t)+\mathbf{H}(q)\mathbf{e}(t),
	\end{align*}
	where $\mathbf{G}(i,j)=G_{ij}$ and $\mathbf{H}=\mbox{diag}(H_1,\cdots, H_J)$. The system matrix $
	\mathbf{G}$ reflects the underlying topology (presence or absence of connections) of the system: $G_{ij}\neq 0$ indicates an edge going from node $j$ to node $i$. Let us further assume that the network is sparse in the number of edges (which is typically the case in practical systems), and that the generating dynamic process is stable. 
	
	Then the model in \eqref{eq:dyn_sys_model_1} may be equivalently described in the prediction error minimization framework \citep{DBLP:journals/corr/abs-1903-06205} as
	\begin{align*}
	w_i(t)=w_i(t|t-1)+e_i(t),
	\end{align*}
	where  \begin{align}
	w_i(t|t-1)=\sum_{j=1}^J\sum_{k=1}^{+\infty} \theta^{(k)}_{ij}q^{-k}w_i(t)
	\label{eq:predictor_full}
	\end{align} 
	is the best one-step predictor of $w_i(t)$ given the past data until time $t-1$, where $\theta_{ij}^{(k)}$ denotes the $k$th impulse response coefficient relating the signal at node $j$ to that at node $i$ \citep{Chiuso2012}. Collecting the signal values at all the $J$ nodes upto time $t-1$ in the vector $\mathbf{a}_i(t)$, and the corresponding impulse response coefficients in the vector $\theta_i$, the signal model at the $i$th node becomes  
		\begin{align*}
		w_i(t)=\mathbf{a}_i(t)^\top\theta_i+e_i(t).
		\end{align*} 
		Since the network is sparse, $\theta_i$ is also sparse (block sparse, more specifically):  $\theta_{ij}^{(k)}=0,\,\forall k$ if $G_{ij}=0$. 
		Thus, learning both the topology and dynamics of the underlying network is equivalent to the estimation of the sparse $\theta_i$ for every node. While different approaches have been pursued to address this problem \citep{Chiuso2012, JAHANDARI2018575,DBLP:journals/corr/abs-1903-06205} they typically involve the estimation of hyperparameters or hyperpriors. As a result, they become demanding both in the amount of data and the computational complexity, specially when one deals with networks of moderate sizes.
		
		Our goal in this work is to develop a method that learns sparse linear dynamic networks in a computationally efficient and hyperparameter-free manner. We do this by building upon an iterative tuning-free covariance-matching based sparse estimation technique known as SPICE (Sparse Iterative Covariance Estimation) \citep{StoicaEtAl2011_spice,Zachariah&Stoica2015_onlinespice} Specifically, the highlights of our method are as follows:
	\begin{itemize}
		\item {\it No hyperparameter-tuning or cross-validation involved} unlike many approaches where the number of hyperparameters scales linearly with the number of nodes in the network 
		\item  {\it Computational efficiency} as a result of the parallel structure and no hyperparameter tuning involved 
			\item {\it Minimal model assumptions} since we do not assume prior knowledge or structure on the predictor parameters 
		\item {\it Accomodates varying noise levels across nodes (Heteroscedastic noise)} since our approach inherently treats every node independently

	\end{itemize}
	
	\subsection{Related Work}
	Various approaches have been considered in the estimation of topology and dynamics in network systems. Frequency-domain methods have been proposed for identification of networks \citep{5406168,6966724} Module-based analysis and characterization of dynamic networks and conditions on identificability under diverse system conditions have been studied \citep{8447494,7798971,7526087,7402842,6426840,VANDENHOF20132994}. Regularized regression approaches which leverage on model sparsity have been explored\citep{JAHANDARI2018575,MATERASSI2013664,doi:10.1111/j.1467-9868.2005.00532.x,Bolstad:2011:CNI:2333135.2333233}. By modelling the transfer function parameters as structured random variables, Bayesian approaches have been also developed \citep{Chiuso2012,Chiuso2015,DBLP:journals/corr/abs-1903-06205,7798971,EVERITT2018144}. 
	Among these, the works of \cite{Chiuso2012} and \cite{DBLP:journals/corr/abs-1903-06205} are the closest in context to our contribution. However, unlike our approach, these works involve computation of hyperparameters along with the use of explicit prior structure. We note that one of our motivations for using the SPICE framework comes from the recent work of \cite{Arun_graphlearn_SPL}, which employs SPICE in estimating static sparse partial correlation graphs from multivariate data.

	\section{The linear dynamic network system model}

		As noted in the Introduction, an equivalent model for \eqref{eq:dyn_sys_model_1} in the prediction error minimization framework is given by
			\begin{align*}
	w_i(t)=w_i(t|t-1)+e_i(t),
	\end{align*}
	where  \begin{align*}
	w_i(t|t-1)=\sum_{j=1, j\neq i}^J\frac{G_{ij}(q)}{H_i(q)}w_j(t)+(1-H_{i}(q)^{-1})w_i(t),
	\end{align*}
	 is the best one-step predictor of $w_i(t)$ given the past data until time $t-1$. Then, by defining the impulse responses
	 \begin{align*}
	 (G_{ij}(q)/H_i(q))\triangleq\sum_{k=1}^{+\infty} \theta^{(k)}_{ij}q^{-k}\\
	 (1-H_i(q)^{-1})\triangleq\sum_{k=1}^{+\infty} \theta^{(k)}_{ii}q^{-k},
	 \end{align*} we arrive at the predictor form in \eqref{eq:predictor_full}. In practice, however, one considers the finite-impulse response (FIR) approximation:
	\begin{align*}
	w_i(t|t-1)=\sum_{j=1}^J\sum_{k=1}^{K} \theta^{(k)}_{ij}q^{-k}w_i(t). 
	\end{align*}
	Notice that the summation over $k$ is now over a finite window of $K$ delays or past samples. The larger the $K$, better would be the model. Let $\mathbf{w}_i^N$ denote the vector containing $N$ consecutive samples of $w_i$ from $t$ to $t+N$. Further, let us define the matrices $A_{j}\in\mathbb{R}^{N\times K}$ for $j=1,\cdots,J$ as
	\begin{align*}
	A_j=\left[\begin{matrix*}
	w_j(t-1)& w_j(t-2)& \cdots w_j(t-K+1)\\
	w_j(t)& w_j(t-2)& \cdots w_j(t-K)\\
	\vdots&\vdots&\vdots&\\
	w_j(t+N-1)& w_j(t+N-2)& \cdots w_j(t+N-K)
	\end{matrix*}\right].
	\end{align*}
	Then, we have that 
	where 
	\begin{align*}
	\mathbf{w}_i^N=\sum_{j=1}^JA_j\theta_{ij}+\mathbf{e}_i^N
	\end{align*}
	where $\theta_{ij}=[\theta_{ij}^{(1)},\cdots,\theta_{ij}^{(K)}]^\top\in\mathbb{R}^{JK}$, and $\mathbf{e}_i^N$ denotes the noise or innovation over the $N$ time samples. Equivalently, by defining $\mathbf{A}_i=[A_i\, \cdots\, A_J]\in\mathbb{R}^{N\times JK}$ and $\theta_i=[\theta_{i1}\,\cdots\,\theta_{iJ}]^\top$, we get the predictor model:
	\begin{align*}
	\mathbf{w}_i^N=\mathbf{A}_i\theta_{i}+\mathbf{e}_i^N,
	\end{align*}
	where $\theta_i$ is sparse, since $\theta_{ij}=0$ when there is no edge from node $j$ to node $i$.
	\section{ Learning sparse linear dynamic networks}
	Given the network system model, there are broadly two ways of approaching the network estimation problem. The first is to consider that the parameter $\theta_i$ is deterministic and unknown. Then, one of the most natural methods would be to solve
	\begin{align}
	\min\sum_{i=1}^J\|\mathbf{w}_i^N-\mathbf{A}_i\theta_i\|_2^2\,\,\mbox{ subject to }\, \|\theta_i\|_0\leq R_i\,\forall i
	\label{eq:l0_norm}
	\end{align}
	where $\|\|_0$ denotes the $\ell_0$ (pseudo) norm which counts the number of non-zeros in the argument. Such an approach requires specification of the upper bounds $R
	_i$, and more importantly is an intractable combinatorial problem []. Hence, various convex relaxations of \eqref{eq:l0_norm} are often used, such as the LASSO (least absolute shrinkage and selection operator) \citep{Tibshirani_LASSO,doi:10.1111/j.1467-9868.2005.00532.x} which solve the following
	\begin{align}
	\min\sum_{i=1}^J\left(\|\mathbf{w}_i^N-\mathbf{A}_i\theta_i\|_2^2+\sum_j\lambda_{ij}\|\theta_{ij}\|_1\right)
	\label{eq:l1_norm}
	\end{align}
	where $\lambda_{ij}$ are the regularization hyperparameters corresponding to the $j$th node for the estimation at the $i$th node. We see that such approaches require cross validation and the number of hyperparameters scale with the number of nodes, if further simplifications such as assuming $\sigma_i^2$ to be the same across nodes are not employed.
	
	The second approach to solving the problem is to assume that $\theta_{i}$ is a stochastic variable with a given prior and to treat the estimation in a completely Bayesian setting with hyperpriors or kernels. In such a case, it is usually assumed that the prior on $\theta_{ij}$ takes the form of kernels which have their own hyperparameters. The optimal estimate is then obtained by maximizing the posterior probability. For certain choice of the priors, the Bayesian estimation becomes identical to solving \eqref{eq:l1_norm}. Nevertheless, one has to still compute for the hyperparameters even in this approach.
	
	An alternative, which we pursue in this work, is to take an intermediate path by assuming that $\theta_i$ for every $i$ is a random variable with expected value $\mathbf{0}$ and a diagonal covariance, prior to observing the data at time $t$. More precisely, we have
	\begin{align*}
	\mbox{E}(\theta_i|t-1)=0,\,\, \mbox{and}\,\, \mbox{Cov}(\theta_i|t-1)=\mbox{diag}(\Pi_i)
	\end{align*}
	where $\Pi_i$ is the vector of variances of different components of $\theta_i$. Then, the optimal estimate maximum aposteriori (MAP) estimate of $\theta_i$ is given by
	\begin{align*}
	\theta_i=(\mathbf{A}_i^\top\mathbf{A}_i+\sigma_i^2\mbox{diag}(\Pi_i))^{-1}\mathbf{A}_i^\top\mathbf{w}^N_i.
	\end{align*}
	Then, the hyperparameters $\Pi_i$ and $\sigma_i^2$ are estimated by matching the posterior covariance of $\mathbf{w}_i^N$ given by
	\begin{align*}
	\mbox{Cov}(\mathbf{w}^N_i|t-1)=\mathbf{A}_i\mbox{diag}(\Pi_i)\mathbf{A}_i^\top+\sigma_i^2\mbox{diag}(\Pi_i),
	\end{align*}
	with the sample covariance $\mathbf{w}^N_i(\mathbf{w}^N_i)^\top$. Such a covariance -matching treatment is well studied in sparse estimation problems. Further, the resulting estimation is known to be equivalent to solving the weighted square root LASSO \citep{StoicaEtAl2014_weightedspice,sqrt_LASSO:}:
	\begin{align}
	\min_{\theta_i}\|\mathbf{w}_i^N-\mathbf{A}_i\theta_i\|_2^2+\sum_j\sum_k\lambda_{jk}|\theta_{ij}^{(k)}|,
	\label{eq:weighted_sqrt_lasso}
	\end{align}
	where $\lambda_{jk}$ is given by the $\ell_2$ norm of the $jk$th column of $\mathbf{A}_j$. This approach of sparse estimation through covariance matching is known as SPICE (Sparse Iterative Covariance Estimation).  
	
	Thus, we advocate the use of the SPICE framework for learning sparse dynamic networks. As a result, our method offers the desirable properties for the  problem: 
	\begin{itemize}
		\item 
		it is free from hyperparameter tuning $-$ the effective hyperparameters are specific to the nodes and can hence accommodate noise levels that vary from one node to the other (heteroscedastic noise),
		\item it does not assume any explicit additional knowledge or prior on the parameters, 
		\item it is computationally efficient (scales as $\mathcal{O}(JK^2N^2)$) and even allows for an online formulation \citep{Zachariah&Stoica2015_onlinespice}, and lastly,
		\item the formulation makes no assumptions on the network being directed or undirected. 
	\end{itemize}
	
	\section{Experiments}
	We next apply our approach to realistic dynamic network data in two experiments. We consider the following data generation strategy for both the experiments. We first randomly generate graphs of size $J=8$ at sparsity level $\rho=0.25$, that is, by considering that only $25\%$ of the total number of possible edges are non-zero in the graph. Having generated the topology, we assign a transfer function over each non-zero edge $(i,j)$ randomly as follows: 
	\begin{itemize}
		\item Experiment 1: $G_{ij}$ and $H_i$s are FIR (finite impulse response) filters of length $K=3$, where the impulse response taps are drawn randomly from the uniform distribution over $[0,1]$ using the functions $drss(\cdot)$ and  that generates a random stable discrete state-space system of a given order.
		\item Experiment 2: $G_{ij}$s and $H_i$s are randomly generated stable rational functions with orders randomly chosen between $1$ to $5$.
	\end{itemize}
	The variances $\sigma_i^2$ are drawn randomly from the uniform distribution over $[0,1]$. For a fixed instance of $\mathbf{G}$, we generate corresponding data as
	\begin{align*}
	\mathbf{w}(t)=[I-\mathbf{G}(q)]^{-1}\mathbf{e}(t)
	\end{align*}
	where we consider only those realizations of transfer functions that result in a stable $[I-\mathbf{G}(q)]^{-1}$. 
	
	We note that while $\theta_i$ estimated by our method will have values very close to zero, they are never usually perfect zeros. This is an observation we share with most sparse estimation techniques. As a result, in order to get exact zeros, one must specify a tolerance or threshold $\delta>0$ such that $\theta_j(hk)$ is set to zero if $|\theta_j(hk)|<\delta$. We note here that the thresholding operation is not an inherent part of our method, rather is typical to all graph/network estimation methods--specially when topology identification is the motive. In our experiments, we set $\delta=10^{-1}$.

	We measure the performance of our approach using three different metrics: True positive rate (TPR) which is the ratio of the edges rightly identified by the toal number of true edges, the false positive rate (FPR) which is the ratio of the edges falsely identified to the ratio of zero-edges or no-edges, and lastly the distance ($\mbox{dis}$)
	defined as
	\begin{align*}
	dis=\sqrt{(FPR)^2+(1-TPR)^2}
	\end{align*}
	which measures the deviation from the ground truth or oracle estimation which produces no false edges and identifies all existing edges.
		\begin{figure}[t]
		\centering\subfigure[]{\includegraphics[width=3in]{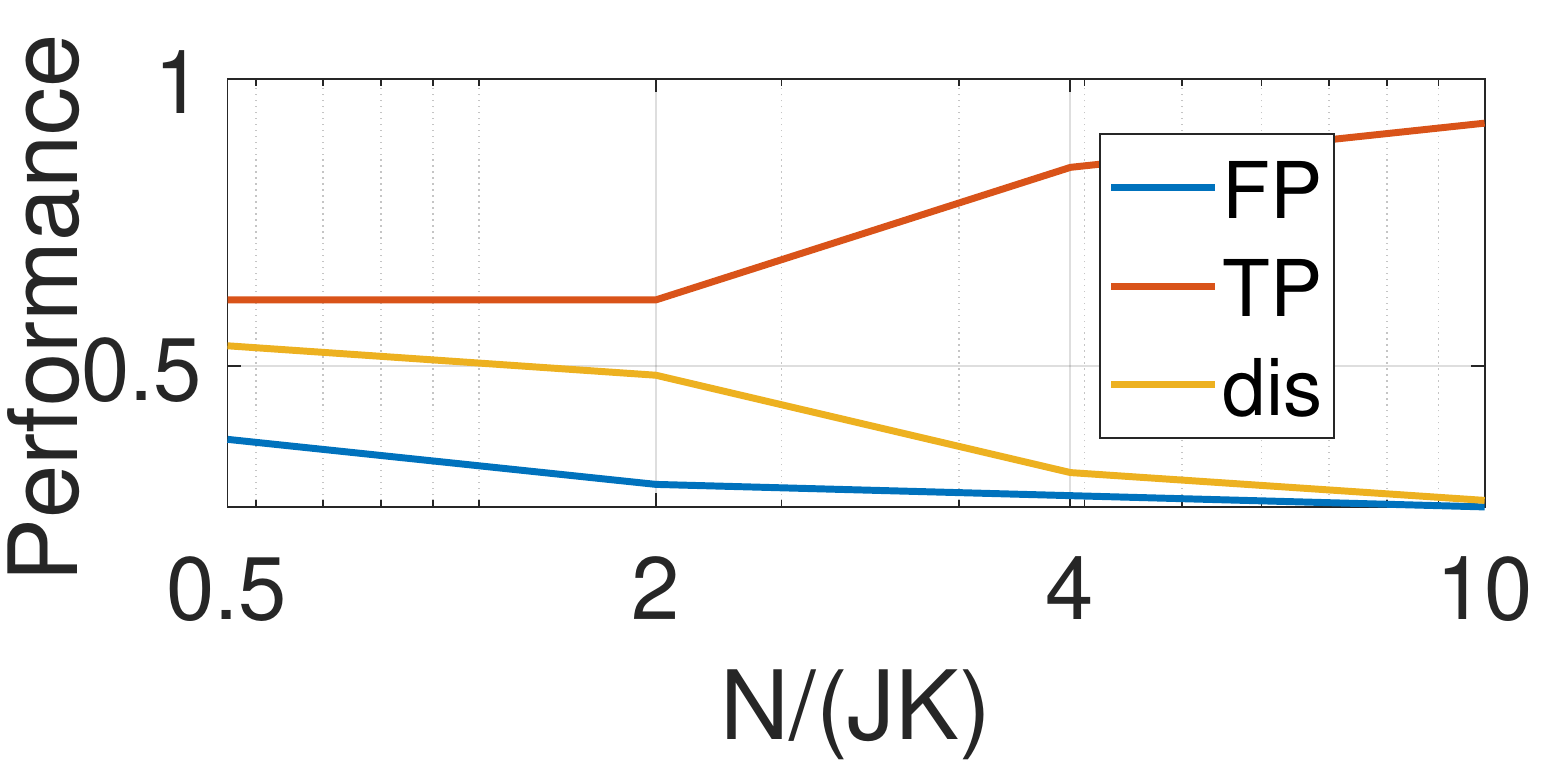}}
		\subfigure[]{\includegraphics[width=3in]{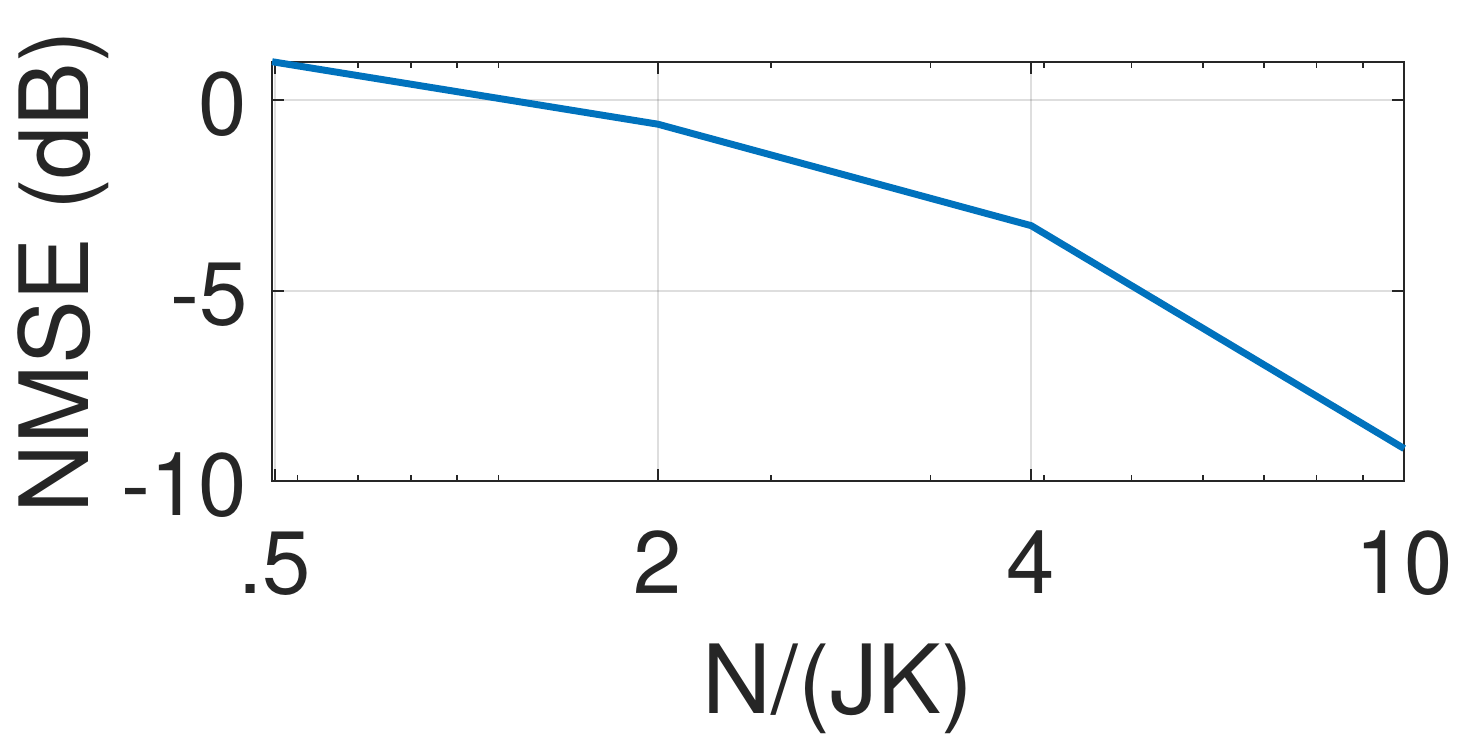}}
		\subfigure[]{\includegraphics[width=2.8in]{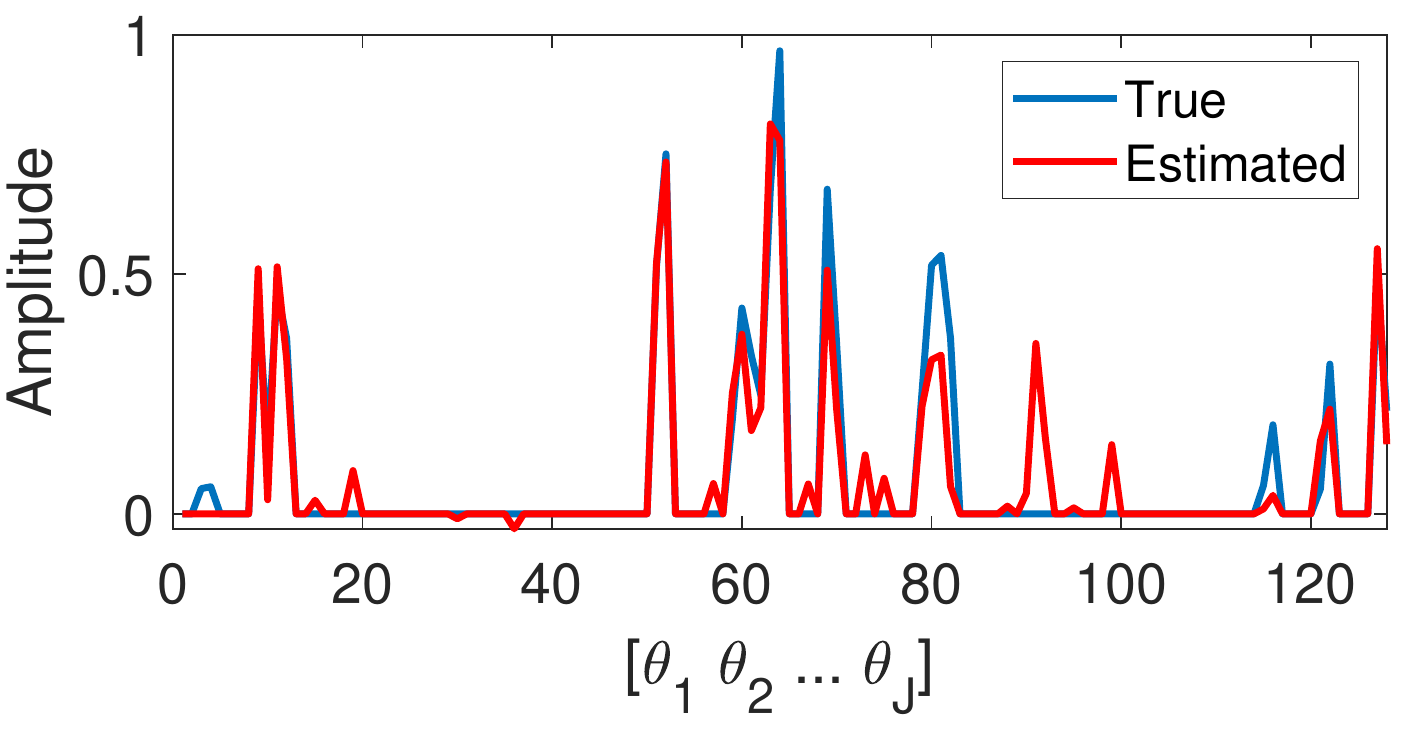}}
		\caption{ Results for Experiment 1 (a) Performance, (b) NMSE in the values of the filter taps (c) An instance of the FIR filter taps estimated.}
		\label{fig:exp1}
	\end{figure}
	In Figure \ref{fig:exp1}(a), we show the performance of our approach in terms of the three metrics as a function of the ratio of the available datasamples $N$ to $JK$ (the dimension of $\theta_i$), averaged over all the $J$ nodes and over 50 Monte-Carlo simulations, for Experiment 1. We observe that as more data becomes available for estimation relative to the number of unknowns, all three metrics improve: TPR gets closer to unity, FPR closer to zero, and $dis$ also tends to zero as $N$ is increased. In order to evaluate the closeness of the estimated FIR taps with that of the ground truth, we evaluate the NMSE in estimation of $\theta_j$ averaged over all nodes, and over the Monte Carlo simulations. The NMSE is shown in Figure \ref{fig:exp1}(b) and shows that NMSE improves progressively as $N$ is increased from very small to large values. Figure \ref{fig:exp1}(c), we show an instance of the true FIR impulse responses used to generate the data and those estimated by our method for $N=1\times JK$. We observe that our estimates for the FIR dynamics almost coincide with the ground-truth.
	
\begin{figure}[t]
		\centering{\includegraphics[width=3in]{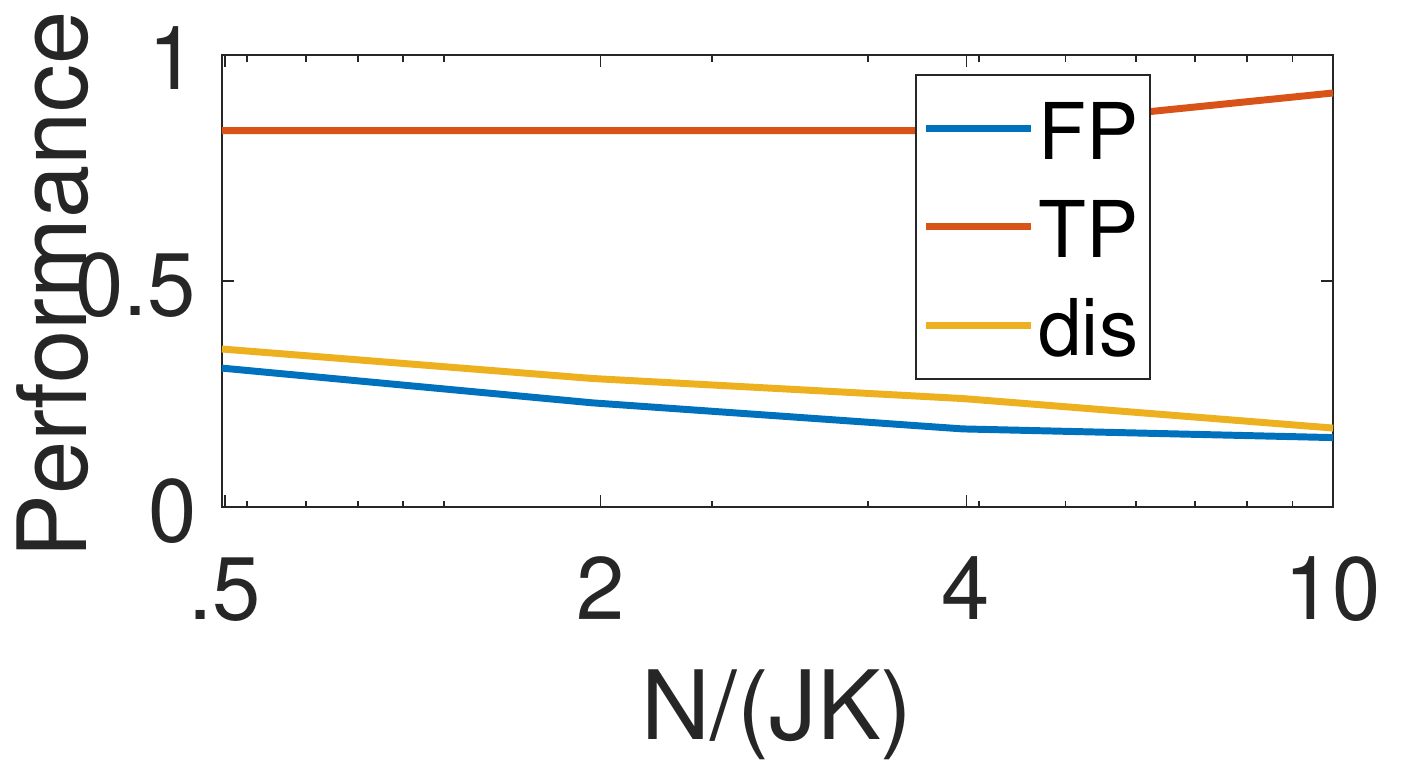}}
		\caption{Performance of our method for Experiment 2.}
		\label{fig:exp2}
	\end{figure}

	In Figure \ref{fig:exp2}, we show the performance of our method for Experiment 2 in terms of the three metrics as $N$ is increased. Once again, we observe the same trend that the estimation improves in terms of all the three metrics as $N$ becomes large.

	\section{Conclusions}
	We proposed a computationally efficient and tuning-free method for learning the underlying topology and dynamics of a sparse linear dynamic system. We achieved this by building upon the hyperparameter-free covariance matching framework. Experiments with realistic linear dynamic systems revealed that our approach results in superior topology identification even with moderate number of available data.
	
\newpage
	\bibliography{ifacconf,refs_graph_learning}             
	
\end{document}